\def\year{2018}\relax
\documentclass[letterpaper]{article} 
\usepackage{aaai18}  
\usepackage{times}  
\usepackage{helvet}  
\usepackage{courier}  
\usepackage{url}  
\usepackage{graphicx}  
\usepackage{amsmath,subfigure,amssymb}
\usepackage{color,tikz,pgfplots}
\usepackage{multirow,paralist}
\usepackage{url}

\newcommand{\newcite}[1]{\citeauthor{#1} (\citeyear{#1})}
\DeclareMathOperator*{\argmax}{arg\,max}
\begin{document}
%
\title{Affective Neural Response Generation}
\author{Nabiha Asghar$^\dagger$, Pascal Poupart$^\dagger$, Jesse Hoey$^\dagger$, Xin Jiang$^\ddagger$, Lili Mou$^\dagger$\\
  $^\dagger$Cheriton School of Computer Science, University of Waterloo, Canada \\
  {\tt \{nasghar,ppoupart,jhoey\}@uwaterloo.ca, \tt doublepower.mou@gmail.com} \\
  $^\ddagger$Noah's Ark Lab, Huawei Technologies, Hong Kong \\
  {\tt jiang.xin@huawei.com} \\}
  
  \maketitle
\begin{abstract}
Existing neural conversational models process natural language primarily on a lexico-syntactic level, thereby ignoring one of the most crucial components of human-to-human dialogue: its affective content. We take a step in this direction by proposing three novel ways to incorporate affective/emotional aspects into long short term memory (LSTM) encoder-decoder neural conversation models: (1) affective word embeddings, which are cognitively engineered, (2) affect-based objective functions that augment the standard cross-entropy loss, and (3) affectively diverse beam search for decoding. Experiments show that these techniques improve the open-domain conversational prowess of encoder-decoder networks by enabling them to produce emotionally rich responses that are more interesting and natural.
\end{abstract}

\section{Introduction}
Human-computer dialogue systems have wide applications ranging from restaurant booking~\cite{task1} to emotional virtual agents~\cite{malhotra2015exploratory}. 
Inspired by the recent success of deep neural networks in natural language processing (NLP) tasks such as language modeling~\cite{LM}, machine translation~\cite{sutskever2014sequence}, and text summarization~\cite{summarization}, the artificial intelligence (AI) community is aggressively exploring the paradigm of neural dialogue generation. 

In a neural network-based dialogue system, discrete words are mapped to real-valued vectors, known as \textit{embeddings}, capturing abstract meanings of  words~\cite{mikolov2013distributed}; then an encoder-decoder framework---with long short term memory (LSTM)-based recurrent neural networks (RNNs)---generates a response conditioned on one or several previous utterances.
Latest advances in this direction have demonstrated its efficacy for both task-oriented dialogue systems~\cite{task1} and open-domain response generation~\cite{shang2015neural,li2016deep,serban2016hierarchical}.

While most of the existing neural conversation models generate syntactically well-formed responses, they are prone to being off-context, short, dull, or vague. Latest efforts to address these issues include diverse decoding \cite{li2016simple,vijayakumar2016diverse}, diversity-promoting objective functions \cite{li2015diversity},
adversarial learning \cite{li2017adversarial}, latent variable modeling for diversity \cite{cao17latent}, human-in-the-loop reinforcement learning
\cite{li2016deep}, 
online active learning \cite{asghar2016online}, 
latent intention modeling \cite{wen2017latentintention}, and content-introducing approaches \cite{mou2016sequence,xing2017topic}. These advancements are promising, but we are still far from our goal of building autonomous neural agents that can consistently carry out interesting human-like conversations.

One shortcoming of these existing open-domain neural conversation models is the lack of \textit{affect} modeling of natural language. These models, when trained over large dialogue datasets, do not capture the emotional states of the two humans interacting in the textual conversation, which are typically manifested through the choice of words, phrases, or emotions. For instance, the attention mechanism in a sequence-to-sequence (Seq2Seq) model can learn syntactic alignment of words within the generated sequences~~\cite{bahdanau2014neural}. Similarly, neural word embedding models like Word2Vec learn word vectors by context, and can preserve low-level word semantics (e.g., ``king''$-$``male''$\approx$``queen''$-$``woman''). However, emotional aspects are not explicitly captured by existing methods.

Our goal is to alleviate this issue in open-domain neural dialogue models by augmenting them with affective intelligence. We do this in three ways. 
\begin{compactenum}
\item We embed words in a 3D affective space by using a cognitively engineered word-level affective dictionary~\cite{warriner2013norms}, where affectively similar constructs are close to one other. In this way, the ensuing neural model is aware of words' emotional features. 
\item We propose to augment the standard cross-entropy loss with affective objectives, so that our neural models are explicitly taught to generate more emotional utterances. 
\item We inject affective diversity into the responses generated by the decoder through \textit{affectively diverse} beam search algorithms, and thus our model actively searches for affective responses during decoding.
\end{compactenum}

We also show that these emotional aspects can be combined to further improve the quality of generated responses in an open-domain dialogue system.

\section{Related Work}
\label{relatedwork}

Affectively cognizant virtual agents are generating interest both in the academia \cite{malhotra2015exploratory} and the industry,\footnote{\url{https://www.ald.softbankrobotics.com/en/robots/pepper}} due to their ability to provide emotional companionship to humans. Endowing text-based dialogue generation systems with emotions is also an active area of research.
Past research has mostly focused on developing hand-crafted speech and text-based features to incorporate emotions in retrieval-based or slot-based spoken dialogue systems \cite{pittermann2010emotion,callejas2011predicting}. 

Despite these, our work is mostly related to two very recent studies: 
\begin{compactitem}
\item Affect Language Model~\cite[Affect-LM]{ghosh2017affectlm} is an LSTM-RNN language model which leverages the Linguistic Inquiry and Word Count \cite[LIWC]{pennebaker2001liwc} text analysis program for affective feature extraction through keyword spotting. Affect-LM considers binary affective features, namely \textit{positive emotion}, \textit{angry}, \textit{sad}, \textit{anxious}, and \textit{negative emotion}; at prediction time, Affect-LM generates sentences conditioned
on the input affect features and a learned parameter of affect strength.

Our work differs from Affect-LM in that we consider affective dialogue systems instead of merely language models, and we have explored more affective aspects including training and decoding.

\item Emotional Chatting Machine~\cite[ECM]{zhou2017ecm} is a Seq2Seq model. It takes as input a prompt and the desired emotion of the response, and then produces a response. It has eight emotion categories, namely \textit{anger}, \textit{disgust}, \textit{fear}, \textit{happiness}, \textit{like}, \textit{sadness}, \textit{surprise}, and \textit{other}. Additionally, ECM contains an internal memory and an external memory. The internal memory models the change of the internal emotion state of the decoder, and therefore encodes how much an emotion has already been expressed. The external memory decides whether to choose an emotional or generic (non-emotional) word at a given step during decoding.

Our approach does not require the input of desired emotion as in \newcite{zhou2017ecm}, which is unrealistic in applications. Instead, we intrinsically model emotion by affective word embeddings as input, as well as objective functions and inference criterion based on these affective embeddings.
\end{compactitem}

\vspace{-.2cm}	

\section{Background}
\label{prelim}

\subsection{Word Embeddings}
In NLP, word embeddings map words (or tokens) to real-valued vectors of fixed dimensionality. In recent years, neural network-based embedding learning has gained tremendous popularity, e.g., Word2Vec \cite{mikolov2013distributed}. They have been shown to boost the accuracy of computational models on various NLP tasks. 

Typically, word embeddings are learned from the co-occurrence statistics of words in large natural language corpora, and the learned embedding vector space has such a property that words sharing similar syntactic and semantic context are close to each other. However, it is known that co-occurrence statistics are insufficient to capture sentiment/emotional features, because words different in sentiment often share context (e.g., ``a \textit{good} book'' vs.~``a \textit{bad} book''). Therefore, we enhance traditional embeddings with a cognitively engineered dictionary, explicitly representing word affective features from several perspectives.

\begin{figure}
\centering
\includegraphics[width=.65\linewidth]{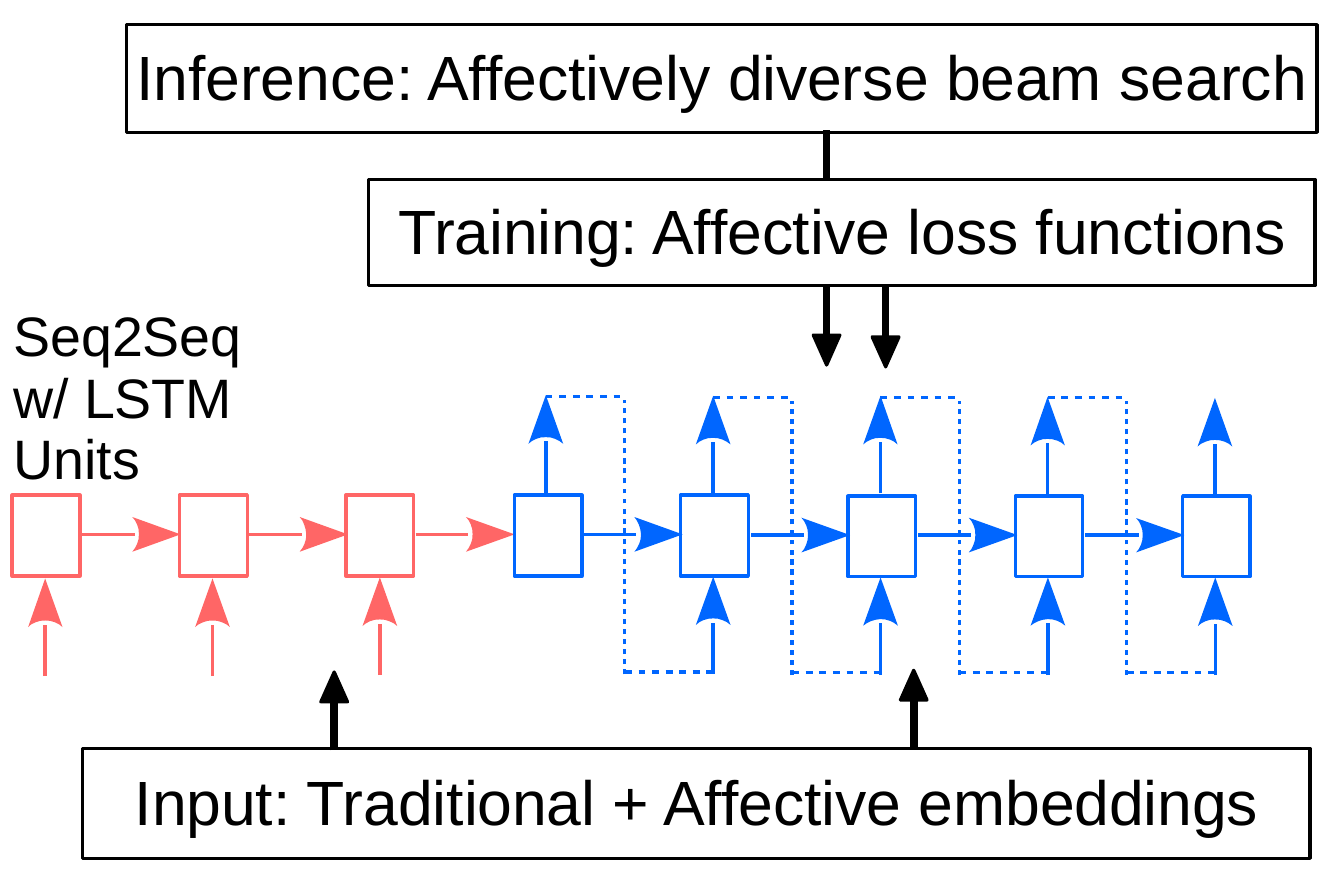}
\vspace{-.3cm}
\caption{Overview of our approach, which is built upon a traditional Seq2Seq model. We propose three affective strategies for the input, training, and inference of Seq2Seq based on a cognitively engineered dictionary with Valence, Arousal, and Dominance (VAD) scores.}
\label{overview}
\vspace{-.3cm}
\end{figure}	

\subsection{Seq2Seq Model}

A sequence-to-sequence (Seq2Seq) model is an encoder-decoder neural framework that maps a variable length input sequence to a variable length output sequence~\cite{sutskever2014sequence}. It consists of an encoder and a decoder, both of which are RNNs (typically with LSTM units). In the context of NLP, the encoder network sequentially accepts the embedding of each word in the input sequence, and encodes the input sentence as a vector. The decoder network takes the vector as input and sequentially generates an output sequence.

Given a message-response pair $(X,Y)$, where $X=x_1, \cdots, x_m$ and $Y=y_1, \cdots, y_n$ are sequences of words/tokens, Seq2Seq models are typically trained with cross entropy loss (XENT):
\begin{equation}
\label{pxent}
L_{\texttt{XENT}}(\theta) = - \log p(Y|X) 
 =  -\sum_{i=1}^{n} \log p(y_i|y_1, \cdots , y_{i-1}, X),
\end{equation}
where $\theta$ denotes model parameters. 
At prediction time, the model generates a response $Y$ to a prompt $X$ by computing
\begin{equation}
\label{pxent2}
 \argmax\nolimits_Y \{ \log p(Y|X) \}
\end{equation}
either greedily or by variants of beam search.



\begin{figure*}[ht!]
\vspace{-.2cm}
\centering
\subfigure[V-A plot.]{
\begin{tikzpicture}[scale=0.65]
\begin{axis}[
legend style={legend pos=north east, font=\small},
xlabel={Valence (V)},
ylabel={Arousal (A)},
xmin=0.5, xmax=9.0,
ymin=0.5, ymax=9.0,
xtick={1.0,3.0,5.0,7.0,9.0},
ytick={1.0,3.0,5.0,7.0,9.0},
ymajorgrids=true,
xmajorgrids=true,
grid style=dashed,
cycle list name=black white
]
\node[label={[magenta]270:{bored}},color=magenta,circle,fill,inner sep=2pt] at (axis cs:2.95,3.65) {};
\node[label={[green]90:{happy}},color=green,circle,fill,inner sep=2pt] at (axis cs:8.47,6.05) {};
\node[label={[green]90:{ecstatic}},circle,color=green,fill,inner sep=2pt] at (axis cs:6.45,6.95) {};
\node[label={[red]90:{angry}},circle,color=red,fill,inner sep=2pt] at (axis cs:2.53,6.2) {};
\node[label={[red]180:{enraged}},circle,color=red,fill,inner sep=2pt] at (axis cs:2.33,5.22) {};
\node[label={[magenta]270:{sad}},circle,color=magenta,fill,inner sep=2pt] at (axis cs:2.1,3.49) {};
\node[label={[magenta]0:{depressed}},circle,color=magenta,fill,inner sep=2pt] at (axis cs:2.27,4.25) {};
\node[label={[green]270:{love}},circle,fill,color=green,inner sep=2pt] at (axis cs:8,5.26) {};
\node[label={[red]180:{hate}},color=red,circle,fill,inner sep=2pt] at (axis cs:1.96,6.26) {};
\node[label={[teal]270:{table}},circle,fill,color=teal,inner sep=2pt] at (axis cs:5.49,3.5) {};
\node[label={[teal]180:{sword}},color=teal,circle,fill,inner sep=2pt] at (axis cs:5.27,5.95) {};
\node[label={[teal]270:{mother}},circle,fill,color=teal,inner sep=2pt] at (axis cs:7.53,4.73) {};
\node[label={[teal]180:{granny}},color=teal,circle,fill,inner sep=2pt] at (axis cs:5.71,2.38) {};

\end{axis}
\end{tikzpicture}
}
\subfigure[V-D plot.]{
\begin{tikzpicture}[scale=0.65]
\begin{axis}[
legend style={legend pos=north east, font=\small},
xlabel={Valence (V)},
ylabel={Dominance (D)},
xmin=0.5, xmax=9.0,
ymin=0.5, ymax=9.0,
xtick={1.0,3.0,5.0,7.0,9.0},
ytick={1.0,3.0,5.0,7.0,9.0},
ymajorgrids=true,
xmajorgrids=true,
grid style=dashed,
cycle list name=black white
]of words (
\node[label={[magenta]90:{bored}},color=magenta,circle,fill,inner sep=2pt] at (axis cs:2.95,4.96) {};
\node[label={[green]90:{happy}},color=green,circle,fill,inner sep=2pt] at (axis cs:8.47,7.21) {};
\node[label={[green]90:{ecstatic}},circle,color=green,fill,inner sep=2pt] at (axis cs:6.45,5.63) {};
\node[label={[red]90:{angry}},circle,color=red,fill,inner sep=2pt] at (axis cs:2.53,4.11) {};
\node[label={[red]-20:{enraged}},circle,color=red,fill,inner sep=2pt] at (axis cs:2.33,4.4) {};
\node[label={[magenta]180:{sad}},circle,color=magenta,fill,inner sep=2pt] at (axis cs:2.1,3.84) {};
\node[label={[magenta]-20:{depressed}},circle,color=magenta,fill,inner sep=2pt] at (axis cs:2.27,3.91) {};
\node[label={[green]270:{love}},circle,fill,color=green,inner sep=2pt] at (axis cs:8,5.92) {};
\node[label={[red]180:{hate}},color=red,circle,fill,inner sep=2pt] at (axis cs:1.96,4.47) {};
\node[label={[teal]215:{table}},circle,fill,color=teal,inner sep=2pt] at (axis cs:5.49,5.84) {};
\node[label={[teal]90:{sword}},color=teal,circle,fill,inner sep=2pt] at (axis cs:5.27,6) {};
\node[label={[teal]90:{mother}},circle,fill,color=teal,inner sep=2pt] at (axis cs:7.53,6.11) {};
\node[label={[teal]-90:{granny}},color=teal,circle,fill,inner sep=2pt] at (axis cs:5.71,5) {};

\end{axis}
\end{tikzpicture}
}
\subfigure[D-A plot.]{
\begin{tikzpicture}[scale=0.65]
\begin{axis}[
legend style={legend pos=north east, font=\small},
xlabel={Dominance (D)},
ylabel={Arousal (A)},
xmin=0.5, xmax=9.0,
ymin=0.5, ymax=9.0,
xtick={1.0,3.0,5.0,7.0,9.0},
ytick={1.0,3.0,5.0,7.0,9.0},
ymajorgrids=true,
xmajorgrids=true,
grid style=dashed,
cycle list name=black white
]
\node[label={[magenta]270:{bored}},color=magenta,circle,fill,inner sep=2pt] at (axis cs:4.96,3.65) {};
\node[label={[green]90:{happy}},color=green,circle,fill,inner sep=2pt] at (axis cs:7.21,6.05) {};
\node[label={[green]90:{ecstatic}},circle,color=green,fill,inner sep=2pt] at (axis cs:5.63,6.95) {};
\node[label={[red]180:{angry}},circle,color=red,fill,inner sep=2pt] at (axis cs:4.11,6.2) {};
\node[label={[red]180:{enraged}},circle,color=red,fill,inner sep=2pt] at (axis cs:4.4,5.22) {};
\node[label={[magenta]180:{sad}},circle,color=magenta,fill,inner sep=2pt] at (axis cs:3.84,3.49) {};
\node[label={[magenta]180:{depressed}},circle,color=magenta,fill,inner sep=2pt] at (axis cs:3.91,4.25) {};
\node[label={[green]0:{love}},circle,fill,color=green,inner sep=2pt] at (axis cs:5.92,5.26) {};
\node[label={[red]90:{hate}},color=red,circle,fill,inner sep=2pt] at (axis cs:4.47,6.26) {};
\node[label={[teal]270:{table}},circle,fill,color=teal,inner sep=2pt] at (axis cs:5.84,3.5) {};
\node[label={[teal]180:{sword}},color=teal,circle,fill,inner sep=2pt] at (axis cs:6,5.95) {};
\node[label={[teal]270:{mother}},circle,fill,color=teal,inner sep=2pt] at (axis cs:6.11,4.73) {};
\node[label={[teal]180:{granny}},color=teal,circle,fill,inner sep=2pt] at (axis cs:5,2.38) {};
\end{axis}
\end{tikzpicture}
}\vspace{-.4cm}
\caption{The relationship between several examples of adjectives, nouns, and verbs on the 3-dimensional VAD scale.}
\vspace{-.3cm}
\label{vadexamples}
\end{figure*}


\section{The Proposed Affective Approaches}
In this section, we propose affective neural dialogue generation, which augments traditional neural conversation models with emotional cognizance. 

Figure~\ref{overview} delineates an overall picture of our approach. We leverage a cognitively engineered dictionary, based on which we propose three strategies for affective dialogue generation, namely affective word embeddings as input, affective training objectives, and affectively diverse beam search.  As will be shown in the experiments, these affective strategies can be combined to further improve Seq2Seq dialogue systems.

\subsection{Affective Word Embeddings}
\label{embedding}

As said, traditional word embeddings trained with co-occurrence statistics are insufficient to capture affect aspects. 
We propose to augment traditional word embeddings with a 3D affective space by using an external cognitively-engineered affective dictionary~\cite{warriner2013norms}.\footnote{Available at \url{http://crr.ugent.be/archives/1003}}

The dictionary we use consists of 13,915 lemmatized English words, each of which is rated on three traditionally accepted continuous and real-valued dimensions of emotion: Valence (V, the pleasantness of a stimulus), Arousal (A, the intensity of emotion produced, or the degree of arousal evoked, by a stimulus), and Dominance (D, the degree of power/control exerted by a stimulus). Sociologists hypothesize that the VAD space (also known as the EPA\footnote{EPA refers to the affective dimensions \textit{Evaluation}, \textit{Potency}, and \textit{Activity}, which are congruent to \textit{Valence}, \textit{Arousal}, and \textit{Dominance}, respectively.} space) structures the semantic relations of linguistic concepts across languages and cultures; it captures almost 70\% of the variance in affective meanings of concepts \cite{osgood1952nature}. VAD ratings have been previously used in sentiment analysis and empathetic tutors, among other affective computing applications \cite{robison2009evaluating,alhothali2015}. To the best of our knowledge, we are the first to introduce VAD to dialogue generation.


The scale of each dimension in the VAD space is from 1 to 9, where a higher value corresponds to higher valence, arousal, or dominance. Thus, V $\simeq 1, 5$ and $9$ corresponds to a word being very negative (e.g., \textit{pedophile}), neutral (e.g., \textit{tablecloth}) and very positive (e.g., \textit{happiness}), respectively. This axis is traditionally used on its own in most sentiment analysis techniques. Similarly, A $\simeq 1, 5$ and $9$ corresponds to a word having very low (e.g., \textit{dull}), moderate (e.g., \textit{watchdog}), and very high (e.g., \textit{insanity}) emotional intensity, respectively. Finally, D $\simeq 1,5$ and $9$ corresponds to a word/stimulus that is very powerless (e.g., \textit{dementia}), neutral (e.g., \textit{waterfall}) and very powerful (e.g., \textit{paradise}), respectively.  The VAD ratings of each word were collected through a survey in \newcite{warriner2013norms} over 1800 participants. We directly take them as the 3-dimensional word-level affective embeddings. 

Some examples of words (including nouns, adjectives, and verbs) and their corresponding VAD values are depicted in Figure~\ref{vadexamples}. For instance, the VAD vectors of the words \textit{ecstatic} and \textit{bored} are $[6.45,6.95,5.63]$ and  $[2.95,3.65,4.96]$, respectively. This means that an average human rates the feeling of being \textit{bored} as more unpleasant (V), less intense (A), and slightly weaker (D), compared with the feeling of being \textit{ecstatic}.  Similarly, the VAD vectors for the nouns \textit{mother} and \textit{granny} are $[7.53,4.73,6.11]$ and $[5.71,2.38,5.00]$, respectively. Thus, mothers are perceived to be more pleasant (V) and more powerful (D) than grannies, and evoke more intense emotions (A). From Figure~\ref{vadexamples}a, we also see some clusters \{\textit{angry, hate, enraged}\} and \{\textit{depressed, sad, bored}\} that are slightly apart on the A axis. Also, the cluster \{\textit{sword, table, granny}\} is fairly neutral on the V axis, compared with the cluster \{\textit{happy, mother, love}\} in Figure~\ref{vadexamples}b. 

For words missing in this dictionary, such as stop words and proper nouns, we set the VAD vector to be the neutral vector $\vec{\eta}=[5, 1, 5]$, because these words are neutral in pleasantness (V) and power (D), and evoke no arousal (A).
Formally, we define ``\textit{word to affective vector}'' (\texttt{W2AV})  as: 
\begin{equation}
\label{w2av}
 \texttt{W2AV}(w) = 
  \begin{cases}
   \text{VAD}(l(w)),        &
\text{if } l(w) \in dict \\
   \vec{\eta} = [5,1,5],        & \text{otherwise}
  \end{cases}
\end{equation}
where $l(w)$ is the lemmatization of the word $w$. 

In this way, words with similar emotional connotations are close together in the affective space, and affectively dissimilar words are far apart from each other. Thus \texttt{W2AV} is suitable for neural processing.

The simplest approach to utilize \texttt{W2AV}, perhaps, is to feed it to a Seq2Seq model as input. Concretely, we concatenate the \texttt{W2AV} embeddings of each word with its traditional word embeddings, the resulting vector being the input to both the encoder and the decoder.

\subsection{Affective Loss Functions}
\label{objectivefunctions}

Equipped with affective vectors, we further design affective training loss functions to explicitly train an affect-aware Seq2Seq conversation model. The philosophy of manipulating loss function is similar to~\newcite{li2015diversity}, but we focus on affective aspects (instead of diversity in general). We have several heuristics as follows.

\subsubsection{Minimizing Affective Dissonance.}
\label{minimizingdeflection}
We start with the simplest approach: maintaining affective consistency between prompts and responses. This heuristic arises from the observation that typical open-domain textual conversations between two humans consist of messages and responses that, in addition to being affectively loaded, are affectively similar to each other. For instance, a friendly message typically elicits a friendly response and provocation usually results in anger or contempt. Assuming that the general affective tone of a conversation does not fluctuate too suddenly and too frequently, we emulate human-human interactions in our model by minimizing the \textit{dissonance} between the prompts and the responses, i.e.， the Euclidean distance between their affective embeddings. This objective allows the model to generate responses that are emotionally aligned with the prompts.

Thus, at the time step $i$, the loss is computed by
\begin{equation}
\begin{split}
&L^i_\texttt{DMIN}(\theta) = -(1-\lambda)\log p(y_i|y_1, \cdots, y_{i-1}, X) \\
&+ \lambda\ {\hat p}(y_i)\bigg\| \sum^{|X|}_{j=1} \frac{\texttt{W2AV}(x_j)}{|X|} - \sum^{i}_{k=1} \frac{\texttt{W2AV}(y_k)}{i}\bigg\|_2
\label{dmin}
\end{split}
\end{equation}
where $\|\cdot\|_2$ denotes $\ell_2$-norm. The first term is the standard XENT loss as in Equation~\ref{pxent}. The sum $\sum_j\frac{\texttt{W2AV}(x_j)}{|X|}$ is the average affect vector of the source sentence, whereas $\sum_k\frac{\texttt{W2AV}(y_k)}{i}$ is the average affect vector of the target sub-sentence generated up to the current time step $i$. 

In other words, we penalize the distance between the average affective embeddings of the source and the target sentences. Notice that this affect distance is not learnable and that selecting a single predicted word makes the model indifferentiable. Therefore, we relax hard prediction of a word by its predicted probability ${\hat p}(y_i)$.
$\lambda$ is a hyperparameter balancing the two factors.

\subsubsection{Maximizing Affective Dissonance.}
\label{maximizingdeflection}
Admittedly, minimizing the affective dissonance does not always make sense while we model a conversation. An over-friendly message from a stranger may elicit anger or disgust from the recipient. Furthermore, from the perspective of training an open-domain conversational agent whose main purpose is to entertain, it is interesting to generate responses that are \textit{not} too affectively aligned with the prompts. Therefore, we design an objective function $L_{\texttt{DMAX}}$ that \textit{maximizes} the dissonance by flipping the sign in the second term in Equation~\ref{dmin}. (Details are not repeated here.)

\subsubsection{Maximizing Affective Content.} Our third heuristic encourages Seq2Seq to generate affective content, but does not specify the polarity of sentiment. This explores the hypothesis that most of the casual human responses are not vague, dull, or emotionally neutral, while the model could choose the appropriate sentiment. Concretely, we maximize the affective content of the model's responses, so that it avoids generating generic responses like ``\textit{yes},'' ``\textit{no},'' ``\textit{I don't know},'' and ``\textit{I'm not sure}.'' That is, at the time step $i$, the loss function is
\begin{equation}
\begin{split}
\label{ac-i}
L^i_\texttt{AC}(\theta) = &-(1-\lambda)\log p(y_i|y_1, \cdots, y_{i-1}, X) \\
&- \lambda\ {\hat p}(y_i)\big\|\texttt{W2AV}(y_i)-\vec{\eta}\big\|_2
\end{split}
\end{equation}
The second term is an affect objective that discourages non-affective words. We penalize the distance between $y_i$'s affective embedding and the affectively neutral vector $\vec{\eta}=[5,1,5]$, so that the model pro-actively chooses emotionally rich words.

\subsection{Affectively Diverse Decoding}
\label{adbs}

In this subsection, we propose affectively diverse decoding that incorporates affect into the decoding process of neural response generation.

Traditionally, beam search (BS) has been widely used for decoding in Seq2Seq models because it provides a tractable approximation of searching an exponentially large solution space. However, in the context of open-domain dialogue generation, BS is known to produce nearly identical samples like ``\textit{This is great.}'' ``\textit{This is great!}'' and ``\textit{This is so great!}'' that lack semantic and/or syntactic diversity \cite{gimpel2013systematic}.

Diverse beam search (DBS) \cite{vijayakumar2016diverse} is a recently proposed variant of BS that explicitly considers diversity during decoding; it has been shown to outperform BS and other diverse decoding techniques in many NLP tasks. 

In the following, we give an overview of BS and DBS, followed by a description of our proposed affective variants of DBS.

\subsubsection{Beam Search (BS).}
BS maintains top-$B$ most likely (sub)sequences, where $B$ is known as the \textit{beam size}.  At each time step $t$, the top-$B$ subsequences at time step $t-1$ are augmented with all possible actions available; then the top-$B$ most likely branches are retained at time $t$, and the rest are pruned.

Let $V$ be the set of vocabulary tokens and let $X$ be the input sequence. 
Ideally, decoding of an entire sequence $\mathbf{y}^*$ is given by
\begin{equation}
\label{decoding_full}
\textbf{y}^* = y^*_1, \cdots , y^*_T = \argmax_{y_1, \cdots , y_T} \bigg[\sum_{t \in T} \log p(y_t|y_{t-1}, \cdots , y_1, X)\bigg]
\end{equation}
where $T$ is the length. 
BS approximates Equation~\ref{decoding_full} by computing and storing only the top-$B$ high scoring (sub)sequences (called beams) at each time step. Let $\textbf{y}_{i,[t-1]}$ denote the $i$th beam stored at time $t-1$, and let $Y_{[t-1]} = \{\textbf{y}_{1,[t-1]}, \cdots, \textbf{y}_{B,[t-1]}\}$ denote the set of beams stored by BS at time $t-1$. 
Then at time $t$, the BS objective is given by:
\begin{equation}
\label{bs}
\begin{split}
Y_{[t]} &= y^{1^*}_{1..t}, \cdots, y^{B^*}_{1..t} \\
&= \argmax_{\substack{\textbf{y}_{1,[t]},\cdots,\textbf{y}_{B,[t]} \\ \in Y_{[t-1]} \times V}} \sum_{b=1}^B \sum_{i=1}^t \log p(y_{b,i}|\textbf{y}_{b,[i-1]}, X)
\end{split}
\end{equation}
\begin{equation*}
\text{subject to}\hspace{2cm} \textbf{y}_{i,[t]} \ne \textbf{y}_{j,[t]} \hspace{5cm}
\end{equation*}
where $Y_{[t-1]} \times V$ is the set of all possible extensions (i.e.,~$V$) based on the beams stored at time $t-1$ (i.e.,~$Y_{[t-1]}$).

\subsubsection{Diverse Beam Search (DBS).}

DBS aims to overcome the diversity problem in BS by incorporating diversity among candidate outputs. 
DBS divides the top-$B$ beams into $G$ groups (each group containing $B'=G/B$ beams) and incorporates diversity between these groups by maximizing the standard likelihood term as well as a dissimilarity metric among the groups.

Concretely, DBS adds to traditional BS (Equation~\ref{bs}) a dissimilarity term $\Delta (Y^1_{[t]}, \cdots , Y^{g-1}_{[t]})[y_t]$ which measures the dissimilarity between group $g$ and previous groups $1, \cdots, g-1$ if token $y_t$ is selected to extend any beam in group $g$. This  is given by
\begin{equation}
\begin{split}
\label{dbs}
Y^g_{[t]} = 
\argmax_{\substack{\textbf{y}^g_{1,[t]},\cdots,\textbf{y}^g_{B',[t]} \\ \in Y^g_{[t-1]} \times V}} \bigg[& \sum_{b=1}^{B'} \sum_{i=1}^t \log p(y^g_{b,i}|\textbf{y}^g_{b,[i-1]}, X) \\
&+ \lambda_g \Delta (Y^1_{[t]}, \cdots , Y^{g-1}_{[t]})[y^g_{b,t}] \bigg]
\end{split}
\end{equation}
\begin{equation*}
\text{subject to }\hspace{1cm} \textbf{y}^g_{i,[t]} \ne \textbf{y}^g_{j,[t]}\hspace{5cm}
\end{equation*}
where $\lambda_g\ge0$ is a hyperparameter controlling the diversity strength. 
Intuitively, DBS modifies the probability in BS as a general scoring function by adding a dissimilar term between a particular sample (i.e., $y_{b,1}^g\cdots y_{b,t}^g$) and samples in other groups (i.e., $Y^1_{[t]}, \cdots , Y^{g-1}_{[t]}$).
We refer readers to \newcite{vijayakumar2016diverse} for the details of DBS. Here, we focus on the dissimilarity metric that can incorporate affective aspects into the decoding phase.

\subsubsection{Affectively Diverse Beam Search (ADBS).}

The dissimilarity metric for DBS can take many forms as used in \newcite{vijayakumar2016diverse}: Hamming diversity that penalizes tokens based on the number of times they are selected in the previous groups, n-gram diversity that discourages repetition of n-grams between groups, and neural-embedding diversity that penalizes words with similar embeddings across groups. Among these, the neural-embedding diversity metric is the most relevant to us. When used with Word2Vec  embeddings, this metric discourages semantically similar words (e.g., synonyms) to be selected across different groups. 

To decode affectively diverse samples, we propose to inject affective dissimilarity across the beam groups based on affective word embeddings. This can be done either at the word level or sentence level. We formalize these notions below. 

$\bullet$ \textit{Word-Level Diversity for ADBS (WL-ADBS)}.
We define the word level affect dissimilarity metric $\Delta_W$ to be
\begin{align}
\label{wladbs}
&\Delta_W (Y^1_{[t]}, \cdots , Y^{g-1}_{[t]})[y^g_{b,t}]  \nonumber  \\
=&-\sum^{g-1}_{j=1}\sum^{B'}_{c=1}\operatorname{sim}\big(\texttt{W2AV}(y^g_{b,t}), \texttt{W2AV}(y^j_{c,t})\big)
\end{align}
\noindent
where $\operatorname{sim}(\cdot)$ denotes a similarity measure between two vectors. In our experiments, we use the cosine similarity function. $y^g_{b,t}$ denotes the token under consideration at the current time step $t$ for beam $b$ in group $g$, and $y^j_{c,t}$ denotes the token chosen for beam $c$ in a previous group $j$ at time $t$. 

Intuitively, this metric computes the cosine similarity of group $g$'s current beam $b$ with all the beams generated in groups $1, \cdots, g-1$. The metric operates at the word level, ensuring that the word affect at time $t$ is diversified across the groups.

$\bullet$ \textit{Sentence-Level Diversity for ADBS (SL-ADBS)}. The word-level metric $\Delta_W$ in Equation \ref{wladbs} does not take into account the overall sentence affect for each group up to time $t$. Thus, as an alternative, we propose a sentence-level affect diversity metric, given by
\begin{align}
\label{sladbs}
&\Delta_S (Y^1_{[t]}, \cdots , Y^{g-1}_{[t]})[y^g_{b,t}] \nonumber \\
=& -\sum^{g-1}_{j=1}\sum^{B'}_{c=1}\operatorname{sim}\big(\Psi(\textbf{y}^g_{b,[t]}), \Psi(\textbf{y}^j_{c,[t]})\big)\\
\text{where}\hspace{1cm} &\Psi(\textbf{y}^k_{i,[t]}) = \sum_{w \in \textbf{y}^k_{i,[t]}} \texttt{W2AV}(w)\hspace{1.5cm}
\end{align}
Here, $\textbf{y}^k_{i,[t]}$ for $k\leq g$ is the $i$th beam in the $k$th group stored at time $t$; $\textbf{y}^g_{b,[t]}$ is the concatenation of $\textbf{y}^g_{b,[t-1]}$ and $y^g_{b,t}$.

The intuition is that, to capture sentence level affect, this metric computes the \textit{cumulative dissimilarity} (given by the function $\Psi(\cdot)$) between the current beam and all the previously generated beams in other groups. This bag-of-affective-words approach is simple but works well in practice, as will be shown in the experimental results.

It should be also noticed that several other beam search-based diverse decoding techniques have been proposed in recent years, including DivMBest \cite{gimpel2013systematic}, MMI objective \cite{li2015diversity}, and segment-by-segment re-ranking \cite{shao2017generating}. All of them use the notion of a \textit{diversity term} within BS; therefore our affect-injecting technique can be potentially used with these algorithms.

\section{Experiments}
\label{experiments}
\subsection{Dataset and Implementation Details}
We evaluated our approach on the Cornell Movie Dialogs Corpus\footnote{The dataset is available at \url{https://www.cs.cornell.edu/~cristian/Cornell_Movie-Dialogs_Corpus.html}} \cite{danescu2011}. It contains ~300k utterance-response pairs, and we kept a vocabulary size of 12,000.
All our model variants used a single-layer LSTM encoder and a single-layer LSTM decoder, each layer containing 1024 cells.
We used Adam~\cite{adam} for optimization with a batch size of 64 and other default hyperparameters. 
Listed as follows are detailed settings for each model.

\begin{compactitem}
\item For the baseline $L_{\texttt{XENT}}$ loss, we used 1024-dimensional Word2Vec embeddings as input and trained the Seq2Seq model for 50 epochs by using Equation~\ref{pxent}.
\item For the affective embeddings as input, we used 1027-dimensional vectors, each a concatenation of 1024-D Word2Vec and 3-D \texttt{W2AV} embeddings. Training was also done for 50 epochs.
\item For affective loss functions ($L_{\texttt{AC}}$, $L_{\texttt{DMIN}}$, and $L_{\texttt{DMAX}}$), we trained the models in two phases. In the first phase, each model was trained using $L_{\texttt{XENT}}$ loss for 40 epochs. In the second phase, each model was fine-tuned using the affective loss functions for 10 epochs. This two-phase approach was adopted because we observed inferior results (syntactical errors in particular) for single-phase training with the full loss functions for 50 epochs.
\item The ADBS decoding was deployed at test time (both  word-level and sentence-level metrics, $\Delta_W$ and $\Delta_S$ in Equations \ref{wladbs} and~\ref{sladbs}, respectively). We set $G=B$ for simplicity, that is, each group contains a single beam. Therefore, diversification among groups in our case was equivalent to diversification among all the beams.
\item The $\lambda$ hyperparameters for $L_{\texttt{DMIN}}$, $L_{\texttt{DMAX}}$, and $L_{\texttt{AC}}$ 
were manually tuned through validation and set to 0.5, 0.4, and 0.5, respectively. For affectively diverse BS, $\lambda$ was set to 0.7 (Equation~\ref{dbs}).
\end{compactitem}

\begin{table}[!t]
	\centering
	\resizebox{\linewidth}{!}{
		\begin{tabular}{|c||c|c|c|} 
			\hline
			\ \ \ \ \multirow{2}{*}{\textbf{Model}}\ \ \ \  & \textbf{Syntactic} & \multirow{2}{*}{\textbf{Natural}} & \textbf{Emotional}\\
			& \textbf{Coherence} & & \textbf{Approp.}\\
			\hline\hline
			Word embed. & 1.48 & 0.69 & 0.41 \\
			\hline
			Word + Affective & \multirow{2}{*}{\textbf{1.71}} & \multirow{2}{*}{\textbf{1.05}} & \multirow{2}{*}{\textbf{1.01}} \\
			embeddings & & &\\
			\hline
		\end{tabular}
	}
	\caption{The effect of affective word embeddings as input.}
	\label{tab:embedding}
\end{table}
\begin{table}[!t]
	\centering
	\resizebox{\linewidth}{!}{
		\begin{tabular}{|c||c|c|c|} 
			\hline
			\multirow{2}{*}{\ \ \ \ \ \ \ \ \textbf{Model}\ \ \ \ \ \ \ \ \ } & \textbf{Syntactic} & \multirow{2}{*}{\textbf{Natural}} & \textbf{Emotional}\\
			& \textbf{Coherence} & & \textbf{Approp.}\\
			\hline\hline
			$L_{\texttt{XENT}}$ & 1.48 & 0.69 & 0.41 \\
			\hline
			$L_{\texttt{DMIN}}$ & \textbf{1.75} & 0.83 & 0.56 \\
			\hline
			$L_{\texttt{DMAX}}$ & 1.74 & 0.85 & 0.58 \\
			\hline
			$L_{\texttt{AC}}$ & 1.71 & \textbf{0.95} & \textbf{0.71} \\
			\hline
		\end{tabular}
	}
	\caption{Effect of affective loss functions.}
	\label{tab:loss}
\end{table}

\subsection{Results}

Recent work employs both automated metrics (e.g., BLEU, ROUGE, and METEOR) and human judgments to evaluate dialogue systems. While automated metrics enable high-throughput evaluation, \newcite{liu2016hownotto} show that these metrics have weak or no correlation with human judgments. It is also unclear how to evaluate affective aspects by automated metrics. Therefore, in this work, we recruited human judges to evaluate our models, following several previous studies~\cite{shang2015neural,mou2016sequence}. 

To evaluate the quality of the generated responses, we had 5 workers to evaluate 100 test samples for each model variant in terms of \textit{syntactic coherence} (Does the response make grammatical sense?), \textit{naturalness} (Could the response have been plausibly produced by a human?) and \textit{emotional appropriateness} (Is the response emotionally suitable for the prompt?). For each axis, the judges were asked to assign each response an integer score of 0 (bad), 1 (satisfactory), or 2 (good). The scores were then averaged for each axis. We also evaluated inter-annotator consistency by Fleiss' $\kappa$ score~\shortcite{fleisskappa}, and obtained a $\kappa$ score of 0.447, interpreted as ``moderate agreement'' among the judges.\footnote{\texttt{https://en.wikipedia.org/wiki/Fleiss\%27\_ kappa}}

\begin{table}[!t]
	\centering
	\resizebox{\linewidth}{!}{
		\begin{tabular}{|c||c|c|c|} 
			\hline
			\multirow{2}{*}{\textbf{Model}} & \textbf{Syntactic} & \textbf{Affective} & \textbf{No. of Emotionally} \\
			& \textbf{Diversity} &  \textbf{Diversity} & \textbf{Approp. Responses} \\
			\hline\hline
			BS & 1.23 &  0.87 & 0.89 \\ 
			\hline
			H-DBS & 1.47 & 0.79 & 0.78 \\
			\hline
			WL-ADBS &  \textbf{1.51} & 1.25 & 1.30\\
			\hline
			SL-ADBS &  1.45 & \textbf{1.31} & \textbf{1.33}\\
			\hline
		\end{tabular}
	}
	\caption{Effect of affectively diverse decoding. H-DBS refers to Hamming-based DBS used in \protect\newcite{vijayakumar2016diverse}. WL-ADBS and SL-ADBS are the proposed word-level and sentence-level 
		affectively diverse beam search, respectively.}
	\label{tab:decoding}
\end{table}

\begin{table*}[!t]
	\centering
	\resizebox{!}{!}{
		\begin{tabular}{|c||c|c|c|} 
			\hline
			{\textbf{Model}} & \textbf{Syntactic Coherence} & {\textbf{Naturalness}} & \textbf{Emotional Appropriateness}\\
			\hline\hline
			\small{Traditional Seq2Seq} & 1.48 & 0.69 & 0.41 \\
			\hline
			\small{Seq2Seq + Affective Embeddings} & 1.71 & 1.05 & 1.01 \\
			\hline
			\small{Seq2Seq + Affective Embeddings \& Loss} & \textbf{1.76} & 1.03 & 1.07 \\
			\hline
			\small{Seq2Seq + Affective Embeddings \&  Loss \&  Decoding} & 1.69 & \textbf{1.09} & \textbf{1.10} \\
			\hline
		\end{tabular}
	}
		\caption{Combining different affective strategies.}
		\label{tab:ablation}
\end{table*}

The evaluation of diversity was conducted separately (i.e., the results in Table~\ref{tab:decoding}). In this experiment, an annotator was presented with top-three decoded responses and was asked to judge
\textit{syntactic diversity} (How syntactically diverse are the five responses?) and \textit{emotional diversity} (How affectively diverse are the five responses?). The rating scale was 0, 1, 2, and 3 with labels bad, satisfactory, good, and very good, respectively. The annotator was also asked to state the number of beams that were emotionally appropriate to the prompt. The scores obtained for each question were averaged. Moreover, we had three annotators in this experiment (fewer than the previous one), as it required more annotations (3 responses for every test sample). The Fleiss' $\kappa$ score for this protocol was 0.471, also signifying ``moderate agreement'' between the judges.

In the following, we first evaluate the performance of three affective strategies individually, namely affective word embeddings as input, affective loss functions, and affectively diverse decoding. Then we show how these strategies can be integrated. 

\subsubsection{Experiment \#1: Affective word embeddings as input.} Table~\ref{tab:embedding} compares Seq2Seq open-domain dialogue systems with and without the affective word embeddings. We see that the cognitively engineered affective embeddings, even with three additional features, largely improve the Seq2Seq model. The improvement is consistent in terms of all three evaluation aspects, and its effect is the most significant in  emotional appropriateness. 

The results show that traditional word embeddings learned end-to-end during training are not sufficient to capture emotional aspects, and that using additional knowledge makes the model more aware of affect.

\subsubsection{Experiment \#2: Affective loss functions.}

We compare in Table~\ref{tab:loss} the proposed loss functions---namely minimizing affective dissonance (\texttt{DMIN}), maximizing affective dissonance (\texttt{DMAX}), and maximizing affective content (\texttt{AC})---with traditional cross-entropy loss (\texttt{XENT}).

As shown in Table~\ref{tab:loss}, \texttt{DMIN} and \texttt{DMAX} yield similar results, both outperforming \texttt{XENT}. Moreover, \texttt{AC} generally outperforms \texttt{DMIN} and \texttt{DMAX} in terms of naturalness and appropriateness. The results imply that forcing the affect vector in either direction (towards or against the previous utterance) helps the model, but its performance is worse than \texttt{AC}. The mediocre performance of $L_{\texttt{DMIN}}$ and $L_{\texttt{DMAX}}$ could be explained by the fact that the relationship between a prompt and a response is not always as simple as minimum or maximum affective dissonance. It is usually much more subtle; therefore it makes more sense to model this relationship through established sociological interaction theories like the Affect Control Theory~\cite{hoey2016affect}. By contrast, the \texttt{AC} loss function encourages affective content without specifying the affect direction; it works well in practice.

Considering both Tables~\ref{tab:embedding} and~\ref{tab:loss}, we further notice that the affective loss function alone is not as effective as affective embeddings. This makes sense because the loss function itself does not provide additional knowledge to the neural network. However, as will be seen in Experiment \#4, these affective aspects are ``orthogonal'' to each other, and can be directly combined.



\begin{table}[!t]
\centering
\resizebox{\linewidth}{!}{
 \begin{tabular}{|l||l|l|} 
 \hline
 \textbf{Message} & \textbf{Model} & \textbf{Response} \\
 \hline\hline
   & Traditional Seq2Seq & {Um, you know\dots} \\
  \cline{2-3}
   & $L_{\texttt{DMIN}}$ & {But you work for the casino?} \\
  \cline{2-3}
 {I wish}  & $L_{\texttt{DMAX}}$ & {You are personally responsible.} \\ 
  \cline{2-3}
  {I was a} & $L_{\texttt{AC}}$ & {It's suicide!} \\
  \cline{2-3}
 {billion-} & $L_{\texttt{XENT}}+$ {Aff. Emb.} & {Don't trouble yourself about it.} \\ 
    \cline{2-3}
 {aire.} & \small{$L_{\texttt{AC}}+$ Aff. Emb.} & {Would it matter in the end.} \\
 \cline{2-3}
  & {$L_{\texttt{AC}}+$ Aff. Emb.} & \multirow{2}{*}{Would it change you?} \\
  & {+ SL-ADBS} &  \\
  \hline\hline
\multirow{7}{*}{I hate}   & Traditional Seq2Seq & {It was supposed to be.} \\
  \cline{2-3}
\multirow{7}{*}{you.}   & $L_{\texttt{DMIN}}$ & {Aww c all guys do\dots} \\ 
  \cline{2-3}
  & $L_{\texttt{DMAX}}$ & {Keeps fairies away!} \\ 
  \cline{2-3}
 & $L_{\texttt{AC}}$ & {I haven't done anything crazy.} \\
  \cline{2-3}
  & $L_{\texttt{XENT}}+$ {Aff. Emb.} & {Don't say that!} \\ 
  \cline{2-3}
  & {$L_{\texttt{AC}}+$ Aff. Emb.} & {I still love you!} \\
 \cline{2-3}
  & {$L_{\texttt{AC}}+$ Aff. Emb.} & \multirow{2}{*}{{I don't want to fight you.}} \\
  & {+ SL-ADBS} & \\
  \hline
\end{tabular}
}
\caption{Examples of the responses generated by the baseline (traditional Seq2Seq) and affective models.}
\label{exampleresponses}
\end{table}

\subsubsection{Experiment \#3: Affectively Diverse Decoding.} We now evaluate our affectively diverse decoding methods. Since evaluating diversity requires multiple decoded utterances for a test sample, we adopted a different evaluation setting as described before. 

Table~\ref{tab:decoding} compares both word-level and sentence-level affectively diverse BS (WL-ADBS and SL-ADBS, respectively) with the original BS and Hamming-based DBS used in \newcite{vijayakumar2016diverse}. 
We see that WL-ADBS and SL-ADBS beat the baselines BS and Hamming-based DBS by a fair margin on affective diversity as well as number of emotionally appropriate responses. SL-ADBS is slightly better than WL-ADBS as expected, since it takes into account the cumulative affect of sentences as opposed to individual words.

\subsubsection{Experiment \#4: Putting them all together.}

We show in Table~\ref{tab:ablation} how the affective word embeddings, loss functions, and decoding methods perform when they are combined. Here, we chose the best variants in the previous individual tests: 
the loss function maximizing affective content ($L_{\texttt{AC}}$) and the sentence level diversity measure (SL-ADBS).

As shown, the performance of our model generally increases when we gradually add new components to it. This confirms that the three affective strategies can be directly combined for further improvement, and we achieve significantly better performance compared with the original Seq2Seq model, especially in terms of emotional appropriateness. 

Note that our setting is different from the Emotional Chat Machine (ECM)~\cite{zhou2017ecm}, the only other known emotion-based neural dialogue system to the best of our knowledge. ECM requires a desired affect category as input, which is unrealistic in applications. It also differs from our experimental setting (and our research goal), making direct comparison infeasible. However, our proposed affective approaches can be potentially integrated to ECM in addition to their manually specified emotion category.

\subsubsection{Case study.}

We present several sample outputs of all the models in Table~\ref{exampleresponses} to give readers a taste of how the responses differ. $L_{\texttt{XENT}}$ responses are generic and non-committal, as expected. $L_{\texttt{DMIN}}$ tries to match the affect of the word \textit{billionaire} with \textit{casino}, $L_{\texttt{DMAX}}$ responds to \textit{hate} with \textit{fairies}, $L_{\texttt{AC}}$ maximizes affective content of the responses with the words \textit{suicide} and \textit{crazy}. $L_{\texttt{XENT}}$ with affective embeddings produces responses with more subtle affective connotations.

\section{Conclusion and Future Work}
\label{conc}
In this work, we address the problem of affective neural dialogue generation, which is useful in applications like emotional conversation partners to humans. We advance the development of affectively cognizant neural encoder-decoder dialogue systems by three affective strategies. We embed linguistic concepts in an affective space with a cognitively engineered dictionary, propose several affect-based heuristic objective functions, and introduce affectively diverse decoding methods.

In the future, we would like to investigate affect-based attention mechanisms for neural conversational models. We would also like to explore affect-based personalization of neural dialogue systems using reinforcement learning.

\section{Acknowledgments}
We thank Marc-Andr\'e Cournoyer for his helpful Github repositories on neural conversation models, and the annotators who helped us evaluate our models.

\small
\bibliography{refs17}
\bibliographystyle{aaai}

\end{document}